%% file: main.tex
\documentclass[10pt,twocolumn,letterpaper]{article}

\usepackage{cvpr}              
\usepackage[accsupp]{axessibility}
\usepackage{ulem}
\usepackage{placeins}
\usepackage{float}
\usepackage{tabularray}


\definecolor{cvprblue}{rgb}{0.21,0.49,0.74}
\usepackage[pagebackref,breaklinks,colorlinks,allcolors=cvprblue]{hyperref}
\usepackage{xcolor} 
\usepackage{colortbl}

\def\paperID{15470} 
\def\confName{CVPR}
\def\confYear{2025}

\title{
Pathways on the Image Manifold:
Image Editing via Video Generation
}

\author{
Noam Rotstein \qquad
Gal Yona \qquad
Daniel Silver \qquad
\\
Roy Velich \qquad
David Bensa\"id  \qquad
Ron Kimmel 
 \\ \\
  Technion - Israel Institute of Technology
}

\begin{document}
\maketitle
\input{sec/0_abstract}    
\input{sec/1_zintro-david}
\input{sec/2_related}
\input{sec/3_method}

\input{sec/4_0_manifold}

\input{sec/4_experiments}
\input{sec/5_limitations}

\input{sec/6_conclusions}
\FloatBarrier
\newpage
{
    \small
    \bibliographystyle{ieeenat_fullname}
    \bibliography{main}
}
\input{sec/7_appendix}


\end{document}

%% file: sec/0_abstract.tex
\begin{abstract}

Recent advances in image editing, driven by image diffusion models, have shown remarkable progress.
However, significant challenges remain, as these models often struggle to follow complex edit instructions accurately and frequently compromise fidelity by altering key elements of the original image.
Simultaneously, video generation has made remarkable strides, with models that effectively function as consistent and continuous world simulators.
In this paper, we propose merging these two fields by utilizing image-to-video models for image editing. 
We reformulate image editing as a temporal process, using pretrained video models to create smooth transitions from the original image to the desired edit. 
This approach traverses the image manifold continuously, ensuring consistent edits while preserving the original image's key aspects.
Our approach achieves state-of-the-art results on text-based image editing, demonstrating significant improvements in both edit accuracy and image preservation.
Visit our \href{https://rotsteinnoam.github.io/Frame2Frame}{project page}.

\vspace{-10pt}

\begin{figure}
\centering
\includegraphics[width=0.98\linewidth]{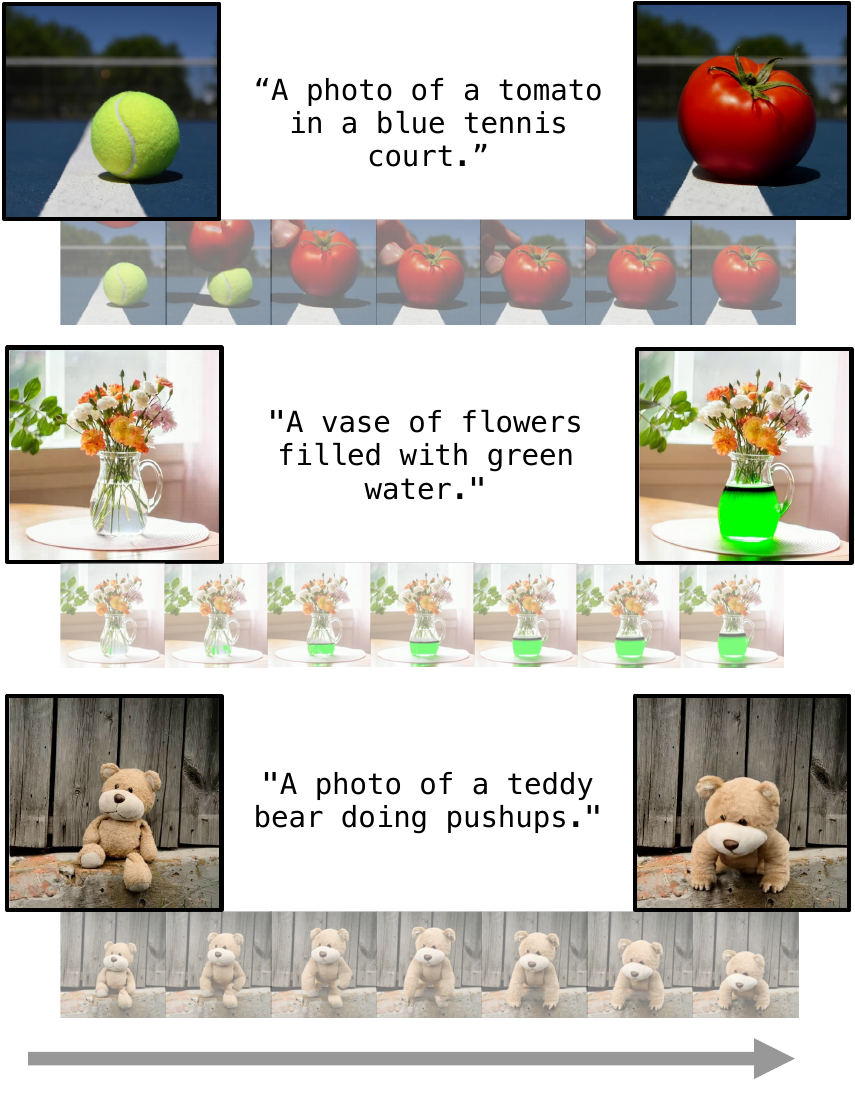}
\vspace{-6pt}
\caption{\textbf{Visualization of Frame2Frame's editing process.}
Temporal progression of our video-based approach.
Starting from the source image (leftmost), frames illustrate the natural evolution toward the target edit (rightmost).
Our method produces temporally coherent intermediate states while preserving fidelity to both the source content and the editing intent.
}
\label{fig:teaser}
\vspace{-0.46cm}

\end{figure}

\end{abstract}

%% file: sec/1_zintro-david.tex
\section{Introduction}

Image editing has witnessed remarkable advancements through deep learning and text-guided diffusion networks.
These developments have set a new benchmark for image manipulations, enhancing both control and quality.
However, current approaches continue to face significant limitations in real-world scenarios. These methods often struggle with two key challenges: achieving precise edits that accurately reflect the intended modifications, and preserving the essential characteristics of the original image content.


State-of-the-art techniques predominantly rely on text-guided diffusion models, which iteratively denoise random latent representations to generate edited images.
Such methods condition the generation process on both the source image—using techniques such as latent inversion \cite{mokady2023null} or model fine-tuning \cite{kawar2023imagic}—and the target edit description.
However, these approaches require the model to generate a single output image that preserves source image fidelity while implementing complex edits, often compromising edit accuracy and content preservation.

In this paper, we propose a paradigm shift in image editing by reformulating it as a video generation task.
Rather than a single-state transition, our approach harnesses temporal coherence: the source image serves as the initial frame of a video that progressively and naturally transforms toward the target edit.
This temporal evolution allows the editing process to unfold through physically plausible intermediate states, providing a continuous path between source and target images, as illustrated in \Cref{fig:teaser} and \Cref{fig:f2f_videos} in the appendix.
This temporal approach leverages the sophisticated world understanding embedded in recent video generation models, which have achieved breakthrough results in temporal coherence and visual quality through training on large-scale internet data \cite{yang2024cogvideox, blattmann2023stable}.
From a geometrical perspective, conventional editing approaches project initial noise onto the natural image manifold, targeting a single point where the image aligns with both the source and the edit request.
In contrast, our approach generates a continuous path  along the manifold between the original and edited image, producing a smooth realistic transition across different image states, as thoroughly discussed in \Cref{sec:manifold}.

We implement the proposed approach through a structured pipeline called \textbf{Frame2Frame} (F2F). First, we transform the edit instruction into a \textit{Temporal Editing Caption} — a scenario describing how the edit should naturally evolve over time—using a pretrained Vision-Language Model (VLM). Next, a state-of-the-art image-to-video model generates a temporally coherent sequence guided by the temporal caption.
Finally, we identify the frame that best realizes the desired edit with the assistance of a VLM.
Extensive experiments demonstrate improvements over existing image-to-image approaches.
We evaluate on TEdBench \cite{kawar2023imagic} and PosEdit, a newly curated dataset derived from UTD-MHAD \cite{chen2015utd}, which focuses on human pose transformations.
PosEdit pairs source images with ground-truth targets of the same subject in different poses, enabling rigorous evaluation of both edit accuracy and source fidelity.
Beyond commonly defined editing tasks, our framework shows promising results in more classical computer vision problems such as de-blurring, de-noising, and relighting by recasting them as temporal progressions, suggesting broader applications for video-based image transformations.

\textbf{Our main contributions include}:
\begin{enumerate}
\item Reformulating image editing as a generative video task--leveraging temporal coherence to create edit paths on the natural image manifold, enabling high-fidelity manipulations while preserving source characteristics.
\item Frame2Frame: an end-to-end framework that realizes the reformulation through three key components: (1) temporal editing captions, (2) generated video-based editing, and (3) automated frame selection.
\item Comprehensive evaluation showing state-of-the-art performance on TEdBench and PosEdit, a new dataset for evaluating human pose edits.
\end{enumerate}

\begin{figure*}
\centering
\includegraphics[width=0.73\textwidth]{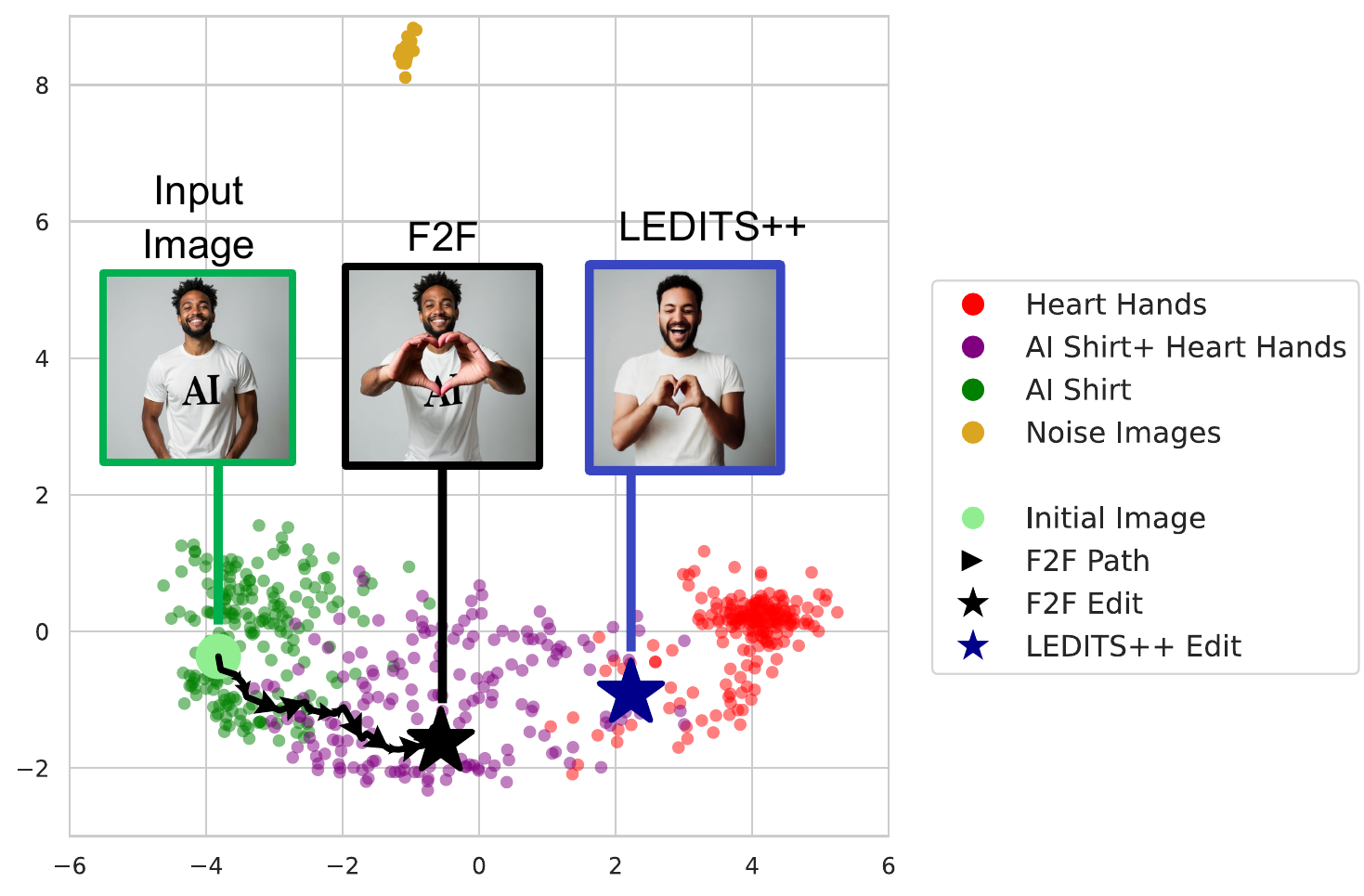}
\vspace{-7pt}
\caption{\textbf{Editing Manifold Pathway.}
Given an input image and target caption "A happy person making a heart shape with their hands", our method generates a continuous path on the natural image manifold.
Each generated frame (indicated by black arrows) represents a plausible intermediate state between the source and target, maintaining temporal consistency throughout the transformation.
As a result, in contrast to the competing approach, F2F achieves the desired edit while preserving the "AI" text on the person’s shirt.
}
\vspace{-8pt}
\label{fig:manifold_walk}
\end{figure*}

%% file: sec/2_related.tex
\section{Related Efforts}
\subsection{Image Editing}
Text-based image editing has advanced significantly with the success of generative diffusion models \cite{ho2020denoising}.
These models, in their text-to-image version, generate images through a denoising diffusion process conditioned on input text, relying on ground truth text-image pairs for training \cite{saharia2022photorealistic, rombach2022high}.
In contrast, image editing often lacks predefined ground truth data for source and target images, posing unique challenges.
This limitation has led researchers to explore diverse editing methodologies \cite{huang2024diffusion}.
For example, SDEdit \cite{meng2021sdedit} injects noise into an image and then denoises it based on an editing target prompt.
Imagic \cite{kawar2023imagic} fine-tunes a text-to-image model on a single image, subsequently interpolating between input and target text embeddings to produce edits.
Other methods first invert the input image into a diffusion model’s latent space \cite{mokady2023null, Huberman-Spiegelglas_2024_CVPR} and then generate the edited image from that latent representation using various techniques for structure preservation and manipulation \cite{hertz2022prompt, parmar2023zero}.
InstructPix2Pix \cite{brooks2023instructpix2pix} synthesizes an editing dataset using the approach in \cite{hertz2022prompt}, filters it, and employs this dataset to train a diffusion model in a supervised fashion.
Paint-by-Inpaint \cite{wasserman2024paint} further explores this supervised approach, generating a real-image dataset for object insertion.


\subsection{Generative Video Models}
Recent years have seen remarkable advancements in video generation, evolving from domain-specific systems \cite{gao2023high} and brief clips to models capable of generating diverse, high-fidelity content.
This progress has stemmed from paradigm shifts and the large-scale expansion of datasets \cite{chen2024panda} and architectures.
Approaches have evolved from recurrent networks \cite{ha2018world, srivastava2015unsupervised} and generative adversarial networks (GANs) \cite{vondrick2016generating, tulyakov2018mocogan, clark2019adversarial} to latent diffusion models (LDMs) \cite{blattmann2023align, blattmann2023stable, danier2024ldmvfi}, which leverage large U-Net or transformer-based architectures alongside vast internet-sourced datasets.
Notable recent efforts include Stable Video Diffusion \cite{blattmann2023stable}, which trains an LDM on large curated datasets, and OpenAI's Sora \cite{liu2024sora},
which achieves impressive results by integrating large architectures with extensive public and private datasets.
These models, often termed "world simulators" due to their emergent understanding of physical dynamics and temporal coherence.
Our work builds upon CogVideoX \cite{yang2024cogvideox},  a transformer-based latent diffusion model that employs a 3D Variational Autoencoder to compress videos across both spatial and temporal dimensions, enhancing coherence.
To improve text-video alignment, CogVideoX also integrates an expert transformer with adaptive LayerNorm, enabling deep fusion of visual and textual modalities.

Close to our work, several recent efforts have utilized the world simulation capabilities of video diffusion models for various computer vision tasks.
Make-A-Video3D \cite{singer2023text} temporally extends static NeRFs using Score Distillation Sampling from video models.
ViVid-1-to-3 \cite{kwak2024vivid} generates images along a camera trajectory around objects to enable novel view synthesis.
PhysDreamer \cite{zhang2025physdreamer} models rigid object properties through 3D Gaussians and material fields, trained via distillation from pretrained video generators.

\subsection{Image Editing and Video}
\vspace{-3pt}
The intersection of image editing and video has received limited attention.
Existing methods focus on sampling pairs of random frames from videos to build image pair datasets that capture the same subject under varying conditions.
For example, AnyDoor \cite{chen2024anydoor} uses the paired frames as an augmentation method, segmenting foreground objects in each frame and assigning one masked object as the target edited appearance of the subject.
MagicFixup \cite{alzayer2024magic} employs these frames to build a dataset focused on refining user-made subject coarse 2D edits.
Recently, drag-based editing approaches \cite{shi2024instadrag, luo2024readout}, used frames extracted from video and computed optical flow to collect a dataset aimed at editing by spatially dragging points within the image.
In contrast to these approaches, which rely on frame sampling for image data collection, we are the first to directly perform image editing using generative video models.

%% file: sec/3_method.tex
\vspace{-3pt}
\section{Frame2Frame}
\vspace{-4pt}
We present Frame2Frame, a framework that reformulates image editing as a temporal transformation process. Our approach leverages video generation models to create natural transitions between source and target images, achieving consistent and realistic edits. The proposed method has three main steps, as illustrated in \Cref{fig:method_image}.
\vspace{-2pt}
\subsection{Temporal Editing Captions}\label{temporal_captions}
\label{sec:tec}
\vspace{-2pt}
Text-based image editing methods typically operate on two inputs: a source image $I_s$ and a target caption $c$, where $c$ specifies the desired modifications to $I_s$.
Our approach differs fundamentally by modeling editing as a temporal process. This requires a novel type of prompt—the \textit{Temporal Editing Caption}, denoted by $\tilde{c}$—that describes the sequential transformation from source to target image.
We construct $\tilde{c}$ by combining information from $I_s$ and $c$ to create a description of how the desired edit unfolds over time.

To automate this process, we leverage recent advances in Vision-Language Models (VLMs) \cite{dai2023instructblipgeneralpurposevisionlanguagemodels, wang2024cogvlmvisualexpertpretrained, wang2023cogvlm, ganz2024question, rotstein2024fusecap}. The VLM, specifically ChatGPT-4o \cite{openai2024chatgpt4o}, is instructed to produce a concise video scenario that highlights how elements within the image change or move over time. The generated caption captures the essential transformations while maintaining a static camera perspective unless movement is necessary.
To improve generation quality, we utilize in-context learning (ICL) \cite{dong2022survey}, providing the VLM with nine exemplar prompt-caption pairs.
The complete prompt template, ICL examples, and an ablation study comparing this approach with directly using the target captions are included in Section \ref{supp:temporal_captions} of the Appendix.

\subsection{Video Generation}
We employ CogVideoX (I2V-5B) \cite{yang2024cogvideox}, a pretrained generative video latent diffusion model utilizing a transformer-based architecture.
Specifically, we use its image-to-video variant, which has been fine-tuned to generate videos starting from an input image $I_s$.
During generation, $I_s$ is encoded and concatenated with noise in latent space, where the model applies a denoising process guided by the temporal caption $\tilde{c}$.
As elaborated in \Cref{sec:manifold}, this conditioning allows generated videos to start from $I_s$ and evolve naturally along the image manifold, maintaining temporal coherence and consistency.
Additionally, the model's transformer architecture enables effective fusion of visual and textual information, allowing precise control over the editing process through our temporal captions. Formally, given the video generator $G$, we define the generation process as:
\vspace{-3pt}
\begin{eqnarray*}
\text{G}(I_s, \tilde{c}) = V = \{f_1, \dots, f_T\}
\end{eqnarray*}
\vspace{-3pt}
where $V$ denotes the generated video with $T$ frames, and $f_t$ represents the frame at timestep $t$.

\subsection{Frame Selection}\label{sec:frame_selection}
\begin{figure}
\centering
\includegraphics[width=\linewidth]{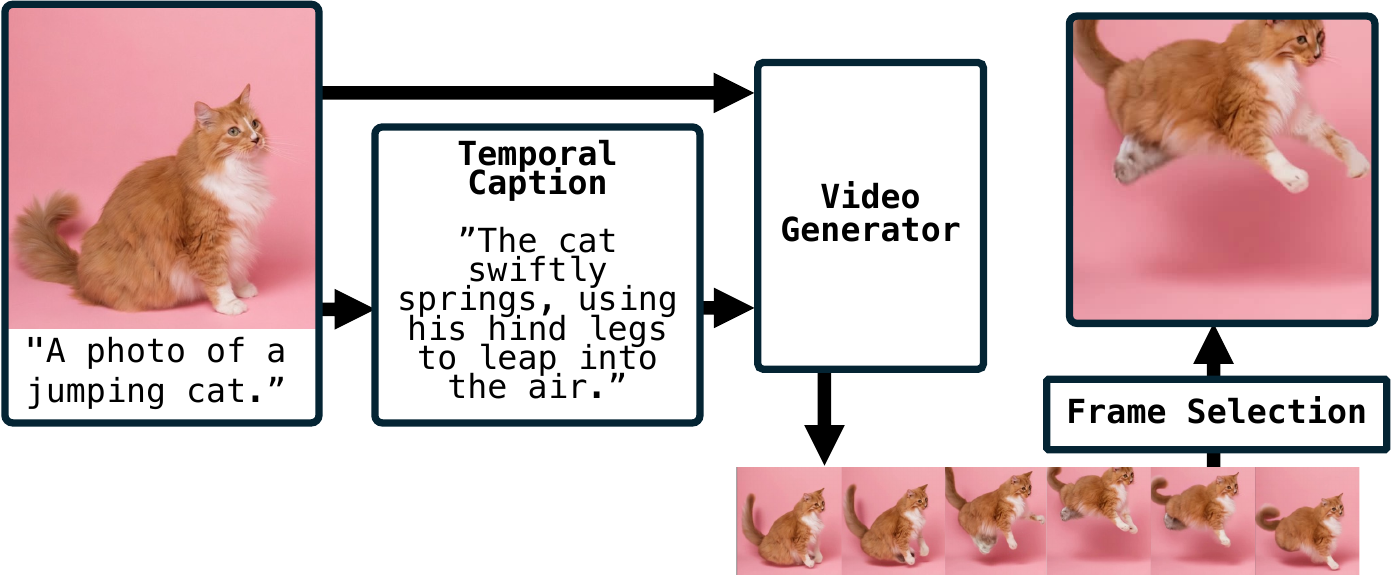}
\vspace{-8pt}
\caption{\textbf{Frame2Frame Overview}. Given a source image and editing prompt, our pipeline proceeds in three steps. First, a Vision-Language Model generates a temporal caption describing the transformation. Next, this caption guides a video generator to create a natural progression of the edit. Finally, our frame selection strategy identifies the optimal frame that best realizes the desired edit, producing the final image of the cat mid-leap.
}
\vspace{-8pt}
\label{fig:method_image}
\end{figure}

We observed that the optimal number of frames required for an edit can vary—small changes may be completed in fewer frames, while more extensive transformations often necessitate additional ones.
Additionally, later frames tend to deviate further from the source image.
Thus, even though $V$ serves as an editing path originating from $I_s$, there is no guarantee that $f_T$ is the optimal edited image in $V$.
For instance, in \Cref{fig:method_image}, the edited image of the cat might be expected to capture the midpoint of its jump, but the last frame shows the jump already completed.
Therefore, we aim to identify the optimal edited frame, denoted $f_{t^*}$, which corresponds to the earliest timestep $t$ that achieves the desired edit.
The transition from the initial frame $f_1$ (or $I_s$) to the final edited frame $f_{t^*}$ motivates our method's name, \textit{Frame2Frame}.

To automate the selection of $t^*$ and avoid manual frame-by-frame review, we employ an automated approach.
After generating the sequence $V$, we sample every fourth frame, imprinting each with a unique identifier and assembling them into an image collage alongside $I_s$.
Inspired by \cite{kim2024image}, which introduces a novel approach to video comprehension by transforming videos into image grids, we use a VLM, specifically GPT-4o, to assist in selecting $t^*$ by providing it with the collage and the editing prompt $c$.
The VLM is tasked with identifying the frame that best fulfills the editing intent, evaluating each frame’s alignment with $c$ and fidelity to $I_s$.
The model is instructed to select the optimal frame with the lowest index that completes the edit.
An ablation study, detailed in Section \ref{supp:frame_seletion} of the appendix, evaluates the effectiveness of our automated approach against the baseline of selecting the final frame of the video.

Note that while we refer to an optimal $t^*$, the definition of the required edit can be subjective and dependent on user preferences.
For example, in \Cref{fig:method_image}, different users may select frames of the cat jumping based on variations in its altitude.
Thus, frame selection could serve as a flexible and customizable advantage in certain scenarios.


%% file: sec/4_0_manifold.tex
\section{Editing Manifold Pathway}\label{sec:manifold}
To illustrate the advantages of our method over conventional image-to-image approaches, we aim to visualize the editing process within the natural image manifold, where realistic images reside.
Given the high dimensionality of this space, we project it into a lower-dimensional representation suitable for interpretation.
To achieve this, we first generate three sets of images, each containing 200 samples, using the text-to-image generator FLUX.1-dev \cite{flux_model}.
The sets are based on the following prompts:
\begin{enumerate} 
    \item \textbf{AI}: \textit{``Full-body portrait of a happy person wearing a shirt with the word 'AI' on it''}.
    \item \textbf{AI + Heart Hands}: \textit{``Full-body portrait of a happy person wearing a shirt with the word 'AI' on it and making a heart shape with their hands''}.
    \item \textbf{Heart Hands}: \textit{``Full-body portrait of a happy person making a heart shape with their hands''}.
\end{enumerate}
We manually filter the outcomes to ensure alignment with descriptions, with examples provided in the supplementary material in Section \ref{supp:manifold}.
Then, we use a CLIP ViT-B/32 model \cite{radford2021learning} to extract a 512-dimensional feature vector for each image.
Using these features, along with 25 feature vectors extracted from random noise images, we perform Principal Component Analysis (PCA) to reduce the dimensionality into a two-dimensional subspace.
The resulting subspace, now suitable for visualization, is depicted in~\Cref{fig:manifold_walk}.
 
As illustrated in the figure, the natural images form a smooth manifold, distinctly separated from the distribution of noise images.
Within this manifold, there is a clear semantic progression: images of people with 'AI' shirts (green cluster) are close to images of people with 'AI' shirts making a heart shape (purple cluster), which are adjacent to images of people only making a heart shape (red cluster).
Thus, transitioning smoothly along the manifold allows a person with an 'AI' shirt to perform a heart shape with their hands while preserving the shirt's text.

Consider an original image from the AI group that we wish to edit using the target prompt: ``A happy person making a heart shape with their hands''.
Current editing methods generate a single image, which may cause an abrupt transition to the red cluster, effectively removing the 'AI' on the shirt.
This behavior is illustrated in the figure, where the edit by LEDITS++ is positioned near the red cluster, and the 'AI' text on the shirt disappears.
In contrast, our method leverages video generation to perform the edit smoothly, moving along the manifold with incremental changes until reaching the required edit in the purple cluster, as indicated by black arrows in \Cref{fig:manifold_walk}.
The temporal editing caption guiding this process is: ``A happy person very slowly raising their hands to form a heart shape''.
This gradual progression lets us reach the purple cluster, resulting in an edited image where the person maintains the `AI' on their shirt while making a heart shape—a faithful preservation of the original image's key attributes.
This experiment illustrates that the proposed paradigm enables smooth traversal across the image manifold, enabling consistent edits while preserving the essential characteristics of the original image.

%% file: sec/4_experiments.tex
\begin{table}
    \centering
    \begin{tabular}{l|cc|c}
        \toprule
        & \multicolumn{2}{c|}{\textbf{Source}} & \textbf{Target} \\
        \raisebox{1.2em}{Model} & 
        \raisebox{1.2em}{LPIPS$_\downarrow$} & 
        \raisebox{1.2em}{CLIP-I$_\uparrow$} & 
        \raisebox{1.2em}{CLIP$_\uparrow$} \\
        \noalign{\vskip -0.6em}
        \midrule
        SDEdit     & 0.30 & 0.85 & 0.60  \\
        Pix2Pix-ZERO         &  0.29 &  0.84 & 0.62 \\
        Imagic            &  0.52 & 0.86 & \textbf{0.63} \\
        LEDITS++            & 0.23 & 0.87 & \textbf{0.63} \\
        FlowEdit & \textbf{0.22} & \textbf{0.89} & 0.61 \\
        \midrule
        F2F    & \textbf{0.22} &  \textbf{0.89} & \textbf{0.63} \\
        \bottomrule
    \end{tabular}
    \caption{
        \textbf{TEdBench Results.} Quantitative comparison on TEdBench benchmark. Source metrics (LPIPS and CLIP-I) measure content preservation, while Target metric (CLIP) measures edit accuracy.
        Our Frame2Frame (F2F) method achieves better or comparable performance across all metrics.
    }
    \label{tbl:tedbench_quant}
\label{tedench_results}
\vspace{-8pt}
\end{table}

\vspace{-3pt}
\section{Experiments}
\vspace{-3pt}
    We evaluate Frame2Frame against state-of-the-art image editing methods, including LEdits++ \cite{leitspp}, SDEdit \cite{sdedit}, Pix2Pix-Zero \cite{pix2pix_zero} Imagic \cite{kawar2023imagic}, and FLUX-based FlowEdit \cite{kulikov2024flowedit}.
    Our evaluation spans two benchmarks: the established TEdBench \cite{kawar2023imagic} for general image editing, and our newly introduced PosEdit, specifically designed for human pose editing.
    Finally, we conduct a human evaluation to assess our method based on real user preferences.
    
    \subsection{Evaluation Protocol}
    We conduct our experiments following a consistent evaluation protocol across all methods and benchmarks.
    Following common practice \cite{kawar2023imagic, leitspp}, for each method and source image, we manually select the best result from fifteen random seeds based on visual quality and edit accuracy, ensuring the same seed set is used across all methods.
    For all methods, we use the default hyperparameters and settings as provided in their official implementations or official Hugging Face repositories
    \footnote{
        \href{https://github.com/ml-research/ledits_pp}{LEdits++}, \href{https://huggingface.co/docs/diffusers/main/en/api/pipelines/pix2pix_zero}{Pix2Pix-Zero},
        \href{https://huggingface.co/docs/diffusers/en/api/pipelines/stable_diffusion/img2img}{SDEdit}
    }.

    \vspace{-4pt}
    \paragraph{Image Preprocessing and Generation Details.}
    We use CogVideoX \cite{yang2024cogvideox} as our video generation backbone, which operates at a fixed resolution of $720 \times 480$ pixels. Both TEdBench and PosEdit benchmarks consist of images with 1:1 aspect ratio. To accommodate CogVideoX's resolution requirements while preserving image content, we first resize all source images to $480 \times 480$ pixels, then horizontally pad them with black pixels to reach the required $720 \times 480$ resolution (adding 120 pixels on each side). After generating the video sequence, we crop the central $480 \times 480$ region of the selected frame to remove the padding and resize it to the final evaluation resolution of $512 \times 512$ pixels, matching the standard resolution used in prior work.
    For video generation, we adopt the default hyperparameters proposed by CogVideoX: a guidance scale of 6, 49 generated frames per sequence, and 50 denoising inference steps.
    This configuration generates videos approximately 6 seconds in duration at a frame rate of 8 frames per second.
\begin{table}
    \centering
        \begin{tabular}{@{}l@{\hspace{3pt}}|@{\hspace{3pt}}c@{\hspace{3pt}}c@{\hspace{3pt}}|@{\hspace{3pt}}c@{\hspace{3pt}}c@{}c@{}}
        \toprule
        & \multicolumn{2}{c|@{\hspace{3pt}}}{\textbf{Source}} & \multicolumn{3}{c@{}}{\textbf{Target}} \\
        \raisebox{0.8em}{Model} & 
        \raisebox{0.8em}{LPIPS$_\downarrow$} & 
        \raisebox{0.8em}{CLIP-I$_\uparrow$} & 
        \raisebox{0.8em}{LPIPS$_\downarrow$} & 
        \raisebox{0.8em}{CLIP-I$_\uparrow$} & 
        \raisebox{0.8em}{CLIP$_\uparrow$} \\
        \noalign{\vskip -0.4em}
        \midrule
        \midrule
        \textcolor{gray}{
        GT (Reference)} &
        \textcolor{gray}{
        0.08} &
        \textcolor{gray}{
        0.91 }
        &
        \textcolor{gray}{0} &
        \textcolor{gray}{1} &
        \textcolor{gray}{0.61} \\
        \midrule
        SDEdit      & 0.39 & 0.61 & 0.39 & 0.64  & 0.57  \\
        Pix2Pix-ZERO          & 0.39 & 0.57 & 0.40 & 0.60  & 0.56 \\
        LEDITS++             & 0.26 & 0.65 & 0.28 & 0.69  & \textbf{0.64}  \\
        \midrule
        F2F         & \textbf{0.14} & \textbf{0.82} & \textbf{0.15} 
        &\textbf{0.84}  & \textbf{0.64} \\
        \bottomrule
    \end{tabular}
    \caption{
    \textbf{PosEdit Results.} 
    Quantitative evaluation on PosEdit.
    Source metrics assess similarity to the original image, while Target metrics include LPIPS and CLIP-I comparisons to the ground-truth target image, along with the CLIP score for edit accuracy.
    “GT” provides ground-truth target image metrics for context. 
    }
    \label{tbl:PosEdit_results}
    \vspace{-8pt}
\end{table}

\begin{figure*}[!htb]
   \centering
     \includegraphics[width=0.9\linewidth]{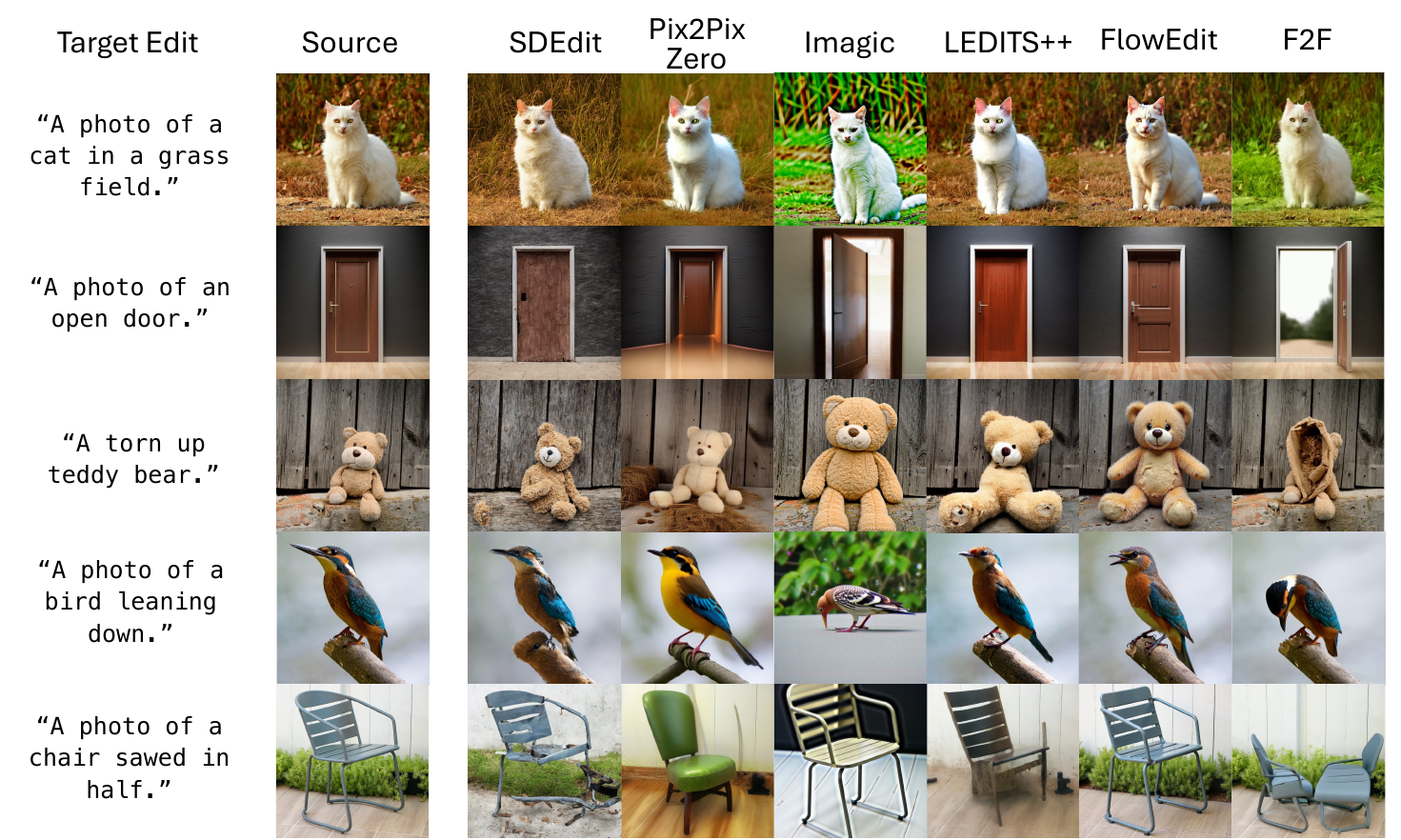}
     \vspace{-2pt}
   \caption{
   \textbf{Qualitative Results on TEdBench.} Comparison with other methods across various editing tasks. Our approach consistently produces edits that better align with the target prompt while preserving the source image's content and structure. For instance, in the teddy bear example, our method uniquely achieves complex structural modifications while maintaining high visual quality.
   }
   \vspace{-8pt}
   \label{fig:tedbench_qualitative}
\end{figure*}

\paragraph{Method-Specific Requirements.}
    Pix2Pix-Zero \cite{pix2pix_zero} requires additional source image descriptions along with the standard inputs.
    We generate these automatically using BLIP-2 \cite{li2023blip}, a state-of-the-art image captioning model, ensuring consistent source descriptions across all experiments.
    For our method, we transform the original editing prompts from both benchmarks into temporal editing captions, as described in  \Cref{sec:tec}.

\vspace{-3pt}
\subsection{TEdBench Evaluation Results}\label{sec:tedbench_results}
\vspace{-3pt}
We quantitatively evaluate our method and baselines on the TEdBench benchmark using three complementary metrics. For each method, we compute the metrics between the source image and its corresponding best edited version. LPIPS \cite{zhang2018perceptual} measures perceptual similarity to assess how much the edit preserves the source image's content, with lower values indicating better preservation. CLIP-I evaluates the similarity between source and edited images in CLIP's \cite{radford2021learning} feature space, where higher values indicate better preservation of semantic content. Finally, CLIP score measures alignment between the edited image and the target prompt, with higher values indicating better adherence to the editing instruction. As shown in \Cref{tbl:tedbench_quant}, our method achieves strong performance across all metrics, demonstrating effective balance between preserving source content and achieving the desired edit.

The qualitative advantages of our method are visually evident in the comparisons presented in \Cref{fig:tedbench_qualitative}.
The figure demonstrates our method's superior performance across diverse editing scenarios, producing results that both faithfully execute the intended edits while maintaining strong alignment with the source image.
Additional examples showcasing these capabilities are provided in the appendix in Section \ref{supp:more_examples}.

\vspace{-3pt}
\subsection{PosEdit Benchmark}\label{sec:posedit}
\vspace{-3pt}

We introduce PosEdit, a benchmark for human pose editing derived from the UTD-MHAD dataset \cite{chen2015utd}.
UTD-MHAD includes RGB videos of 8 subjects (4 females and 4 males) performing predefined actions in a controlled indoor environment. We carefully curated 58 editing tasks encompassing 8 distinct action categories, ranging from simple poses like a raised hand to complex athletic poses such as basketball shooting and lunging.

The proposed benchmark specifically focuses on human pose manipulation.
Unlike TEdBench, each editing task in PosEdit includes a ground truth frame extracted from the same subject in the target pose.
The availability of reference images allows for a more comprehensive evaluation of both editing accuracy and identity preservation.
In the dataset, each editing task consists of two images: a source image showing the subject in a neutral standing pose with arms relaxed at their sides, and a ground-truth edited target image capturing the subject performing in a specific pose. The benchmark also provides a prompt for each task that describes the target pose the subject should achieve (e.g., "A person in a basketball shooting posture.").

\begin{figure}
\centering
\includegraphics[width=\linewidth]{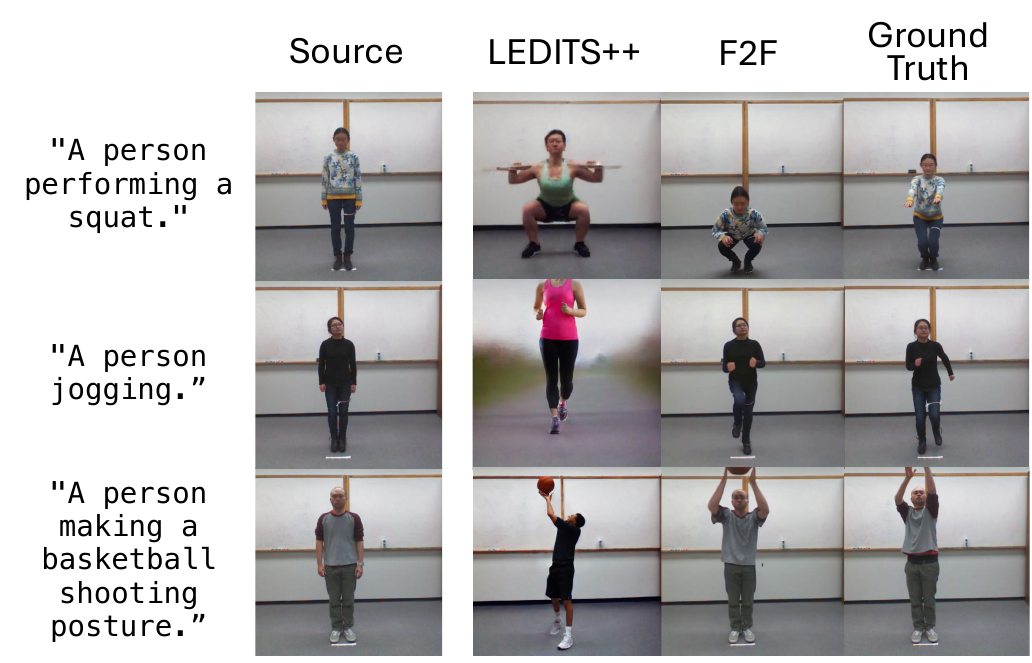}
\vspace{-8pt}
\caption{\textbf{Qualitative Results on PosEdit.} Comparison between our Frame2Frame method and LEDITS++ on human motion editing tasks. For each example, we show the source image, edited results from both methods, and the ground-truth target image. Our method better preserves subject identity while achieving more natural pose transitions.
The evaluation metrics for each image are provided in Section \ref{supp:posedit} of the appendix.
}
\vspace{-10pt}
\label{fig:posedit_qualitative}
\end{figure}

\vspace{-6pt}
\paragraph{Benchmark Evaluation.}
    The evaluation on PosEdit follows the same protocol established for TEdBench: generating multiple variants using nine different seeds and manually selecting the best result for each method. However, PosEdit's ground-truth target images enable additional evaluation metrics. Beyond measuring preservation of source content (via LPIPS and CLIP-I between source and edited images) and edit accuracy (via CLIP score with the prompt), we compute LPIPS and CLIP-I metrics between the edited image and its corresponding ground-truth target. This enhanced evaluation directly assesses both the accuracy of the transformation and preservation of identity features.
    Additionally, we include the evaluation metrics of the ground-truth images as a reference to the rest of the results.

    \Cref{tbl:PosEdit_results} demonstrates that our method consistently outperforms competitive methods across all metrics, with particularly strong performance in similarity measures with the target ground-truth image. This advantage highlights our method's superior ability to preserve key features while achieving the desired edit, as illustrated by the qualitative comparisons in \Cref{fig:posedit_qualitative}. The figure shows that our method generates more natural pose transitions while maintaining crucial identity attributes such as facial features, body proportions, and clothing details.
    
    
    
    
    Two important implementation notes:
    First, for Pix2Pix-Zero, which requires source image descriptions, we use the same static prompt across all tasks: "\textit{A person standing naturally with his arms relaxed at his sides}." Second, we exclude Imagic \cite{kawar2023imagic} from this comparison as their official implementation is not publicly available and their results are only reported on TEdBench.


\subsection{Human Evaluation Survey}\label{sec:survey}
To further evaluate our method, we conducted a human survey.
As indicated in \Cref{tbl:tedbench_quant}, LEDITS++ emerges as the most competitive method compared to ours, making it a natural choice for comparison.
Each participant was presented with 20 randomly sampled TEdBench samples, including the source image, the target edit prompt, and the corresponding edited outputs from both methods.
Inspired by the methodology proposed by \cite{zhang2024magicbrush}, participants evaluated each comparison based on two criteria: (1) edit accuracy relative to the prompt, and (2) edit quality—defined as the preservation of visual fidelity to the source image, seamless integration of edited elements, and the overall natural appearance of modifications.
We collected responses from 59 randomly selected online evaluators.
To ensure an unbiased assessment, evaluators were not informed of the study’s objectives.
The results were quantified using two metrics: (i) overall global preference, expressed as a percentage, and (ii) aggregated per-image preference, where a tie resulted in each method receiving 0.5 points.

\begin{table}
    \centering
    \begin{tabular}{@{}l|c@{\hspace{10pt}}c|c@{\hspace{10pt}}c@{}}
        \toprule
        & \multicolumn{2}{c|}{\textbf{Edit Accuracy}} & \multicolumn{2}{c}{\textbf{Edit Quality}} \\
        \raisebox{0.8em}{Method} & 
        \raisebox{0.8em}{Overall} & 
        \raisebox{0.8em}{Per-Image} & 
        \raisebox{0.8em}{Overall} & 
        \raisebox{0.8em}{Per-Image} \\
        \noalign{\vskip -0.4em}
        \midrule
        F2F       & \textbf{54.1\%} & \textbf{53.0\%} & \textbf{65.6\%} & \textbf{67.0\%}  \\
        LEDITS++  & 45.9\% & 47.0\% & 34.4\% & 33.0\%  \\
        \bottomrule
    \end{tabular}
    \vspace{-2pt}
    \caption{
    \textbf{Human Survey Results.} 
    Human evaluation on TEdBench shows that F2F surpasses LEDITS++ in edit accuracy while offering a significant advantage in preserving the original image.
    }
    \vspace{-9pt}
    \label{tbl:survey_results}
\end{table}

Our results, summarized in \Cref{tbl:survey_results}, demonstrate that F2F outperforms LEdits++ on both metrics.
Specifically:
For edit quality, F2F achieved a global preference score of 53\%, slightly surpassing the 47\% obtained by LEdits++.
In terms of edit accuracy, F2F achieved a 65.6\% global preference compared to LEdits++'s 34.4\%.
The per-image preference results mirrored this trend, indicating robustness across individual examples and no significant influence of outliers.
These findings reinforce our claim that smooth temporal editing—using video as a medium—preserves essential scene characteristics while successfully performing the edit.
Our results suggest that the gap between F2F and LEdits++ in source preservation is larger than reflected by the LPIPS scores in \Cref{tbl:tedbench_quant}.
The full survey details are provided in the supplementary materials in Section \ref{supp:human_survey}.

\begin{figure*}[!htb]
\centering
  \includegraphics[width=1.0\linewidth]{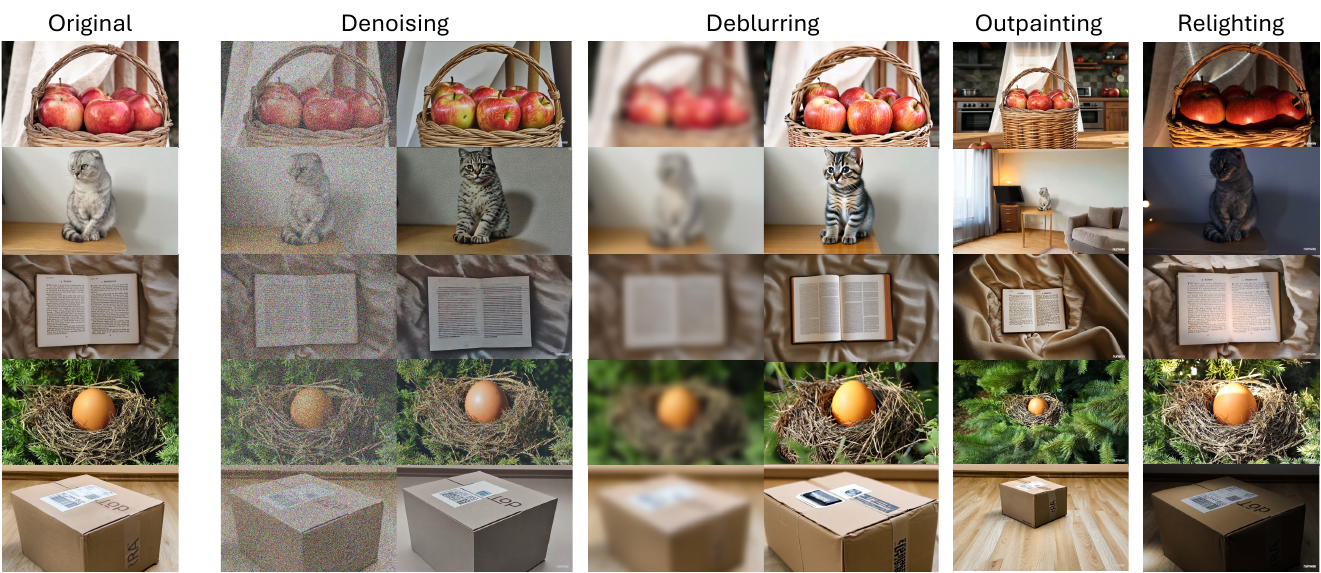}
\caption{
\textbf{Additional Vision Tasks. }
Qualitative results of our image-to-video-to-image editing approach on selected traditional tasks.
}
\label{fig:generatior_example}
\end{figure*}

\subsection{Additional Vision Tasks}\label{sec:silver_tasks}
We demonstrate our framework's applicability beyond traditional editing by applying it to fundamental image manipulation tasks: denoising, deblurring, outpainting, and relighting.
For these experiments, we utilize Runway Gen-3 \cite{runway-gen1} as our video generation backbone, as it showed superior performance on these specific tasks compared to other generative video models.
Since these tasks serve as extensions and potential applications of our method, we focus on qualitative results, leaving quantitative evaluation for future work.
Our results, shown in \Cref{fig:generatior_example}, demonstrate strong performance across all tasks.
We associate this success to the natural alignment between these operations and common video scenarios: deblurring maps to camera focusing, outpainting to camera motion, and relighting to time-lapse lighting changes.
This alignment enables our method to leverage the video model's learned temporal dynamics, achieving high-quality results without task-specific training.
Full prompts are provided in Section \ref{supp:vision_tasks} of the supplementary material.

\vspace{8pt}
\noindent
\textbf{Extended Results.} Additional examples illustrating our method's editing performance across diverse backgrounds are provided in Section \ref{supp:more_ablations} of the Appendix.

%% file: sec/5_limitations.tex
\vspace{-3pt}
\section{Limitations}\label{sec:limitations}
While our approach addresses key shortcomings in current image editing methods, it also introduces unique challenges.
For example, natural camera motion sometimes appears in video sequences, which, when replicated in generated content, can lead to unintended perspective shifts.
As with all generative models, video models are trained on specific data domains, making it challenging to produce results that deviate significantly from the model's training data, which predominantly includes real-world transformations. Despite these challenges, F2F demonstrates success in several cases. For example, in the middle row of \Cref{fig:teaser}, the vase is "magically" filled with green water, showcasing the model's ability to perform imaginative edits, with additional examples discussed in Section \ref{supp:more_ablations} of the supplementary. 
Additionally, our method is computationally intensive, as transforming an image into a video sequence is resource-heavy and often slower than other image editing methods.
However, video generation efficiency is advancing rapidly, as demonstrated by models like Runway Turbo\footnote{\href{https://runwayml.com/news/introducing-the-runway-api}{Runway Turbo API}} and LTX-Video \cite{hacohen2024ltx}, which can execute the process in seconds, making this approach increasingly less resource-intensive.  Additionally, optimizing the number of frames per edit—currently fixed at 49 for CogVideoX—could enable faster editing possibilities.

%% file: sec/6_conclusions.tex
\section{Conclusions}
\vspace{-2pt}
We introduced Frame2Frame, a novel approach that reformulates image editing through video generation. By leveraging video models' inherent understanding of temporal transformations, our method achieves state-of-the-art editing results while maintaining high fidelity to source images. We demonstrated our framework's effectiveness on standard benchmarks, introduced PosEdit for human pose editing, and showed promising results on more classical vision tasks. As video generation technology advances, we expect our approach to enable increasingly sophisticated image manipulations while maintaining natural and physically plausible results.

\vspace{-2pt}
\paragraph{Future Research.}
Our work opens several promising directions.
First, fine-tuning video generators specifically for image editing presents an exciting opportunity, including straightforward solutions like enforcing static camera scenarios or using datasets curated for editing tasks, alongside more complex approaches yet to emerge.
Secondly, a key direction could involve reducing the overhead of full video generation while preserving the benefits of gradual, temporally coherent transformations, potentially enhancing both the efficiency and speed of editing.
The speed can be affected both in terms of per-frame average generation speed or the number of frames generated per edit, which is currently fixed at 49 for CogVideoX.




%% file: sec/7_appendix.tex
\clearpage
\appendix
\renewcommand{\thefigure}{S\arabic{figure}}

\renewcommand{\thetable}{S\arabic{table}}


\twocolumn[
\begin{@twocolumnfalse}
\begin{center}
    {\LARGE \textbf{Appendix}} 
\end{center}
\end{@twocolumnfalse}
]

\section{Additional Editing Examples}\label{supp:more_examples}
\begin{figure}[H]
\centering
\includegraphics[width=1.03\linewidth]{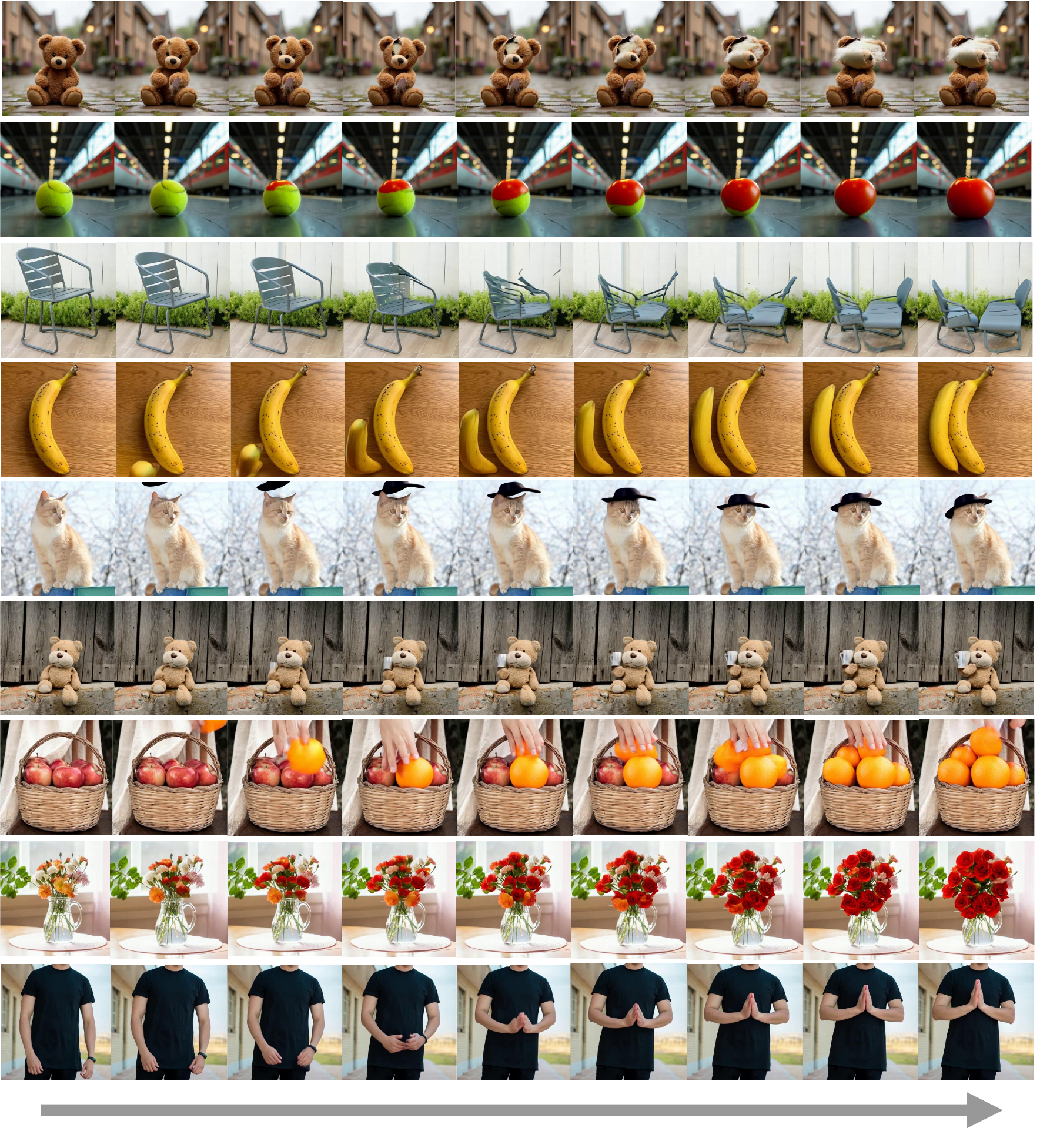}
\caption{\textbf{Generated Video Examples.} Additional video sequences illustrating the temporal evolution from the source image to the target edit.}
\label{fig:f2f_videos}
\end{figure}

We provide additional qualitative examples of the editing results on the TEdBench benchmark, generated based on the experimental setup detailed in  \Cref{sec:tedbench_results}.
These examples further supplement those presented in Figures \ref{fig:teaser} and \ref{fig:tedbench_qualitative}.
Each example is accompanied by the temporal editing caption used to perform the edit, which was generated using the method described in \Cref{sec:tec}.
Furthermore, in \Cref{fig:f2f_videos}, we supplement \Cref{fig:teaser} with additional video examples generated by our method, illustrating the transition from the source image (left) to the target edit (right).

\begin{figure}[H]
\centering
\includegraphics[width=0.95\linewidth]{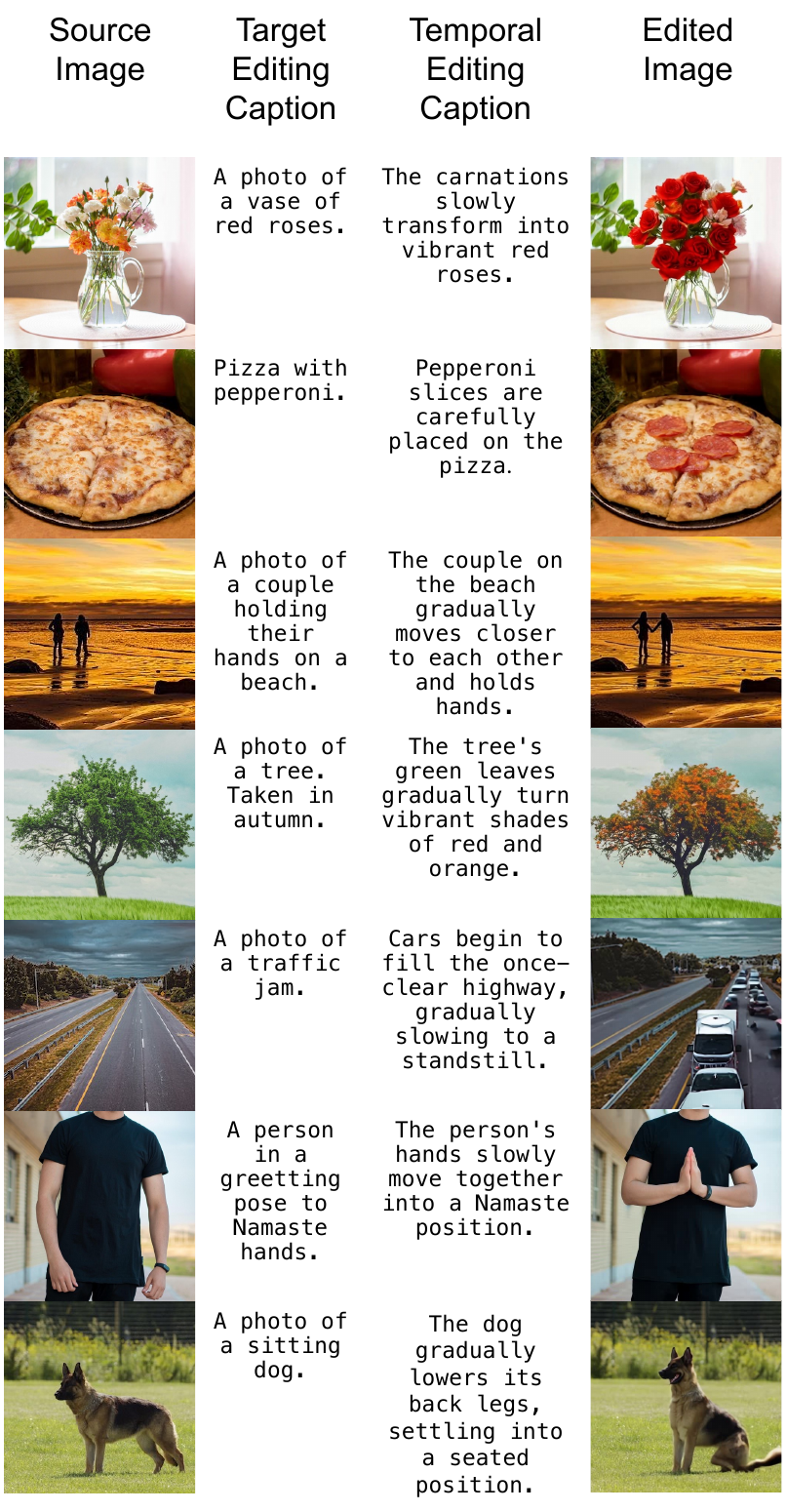}
\caption{\textbf{TEdBench Editing Examples.}}
\label{fig:f2f_examples_1}
\end{figure}

\begin{figure}[H]
\centering
\includegraphics[width=1.04\linewidth]{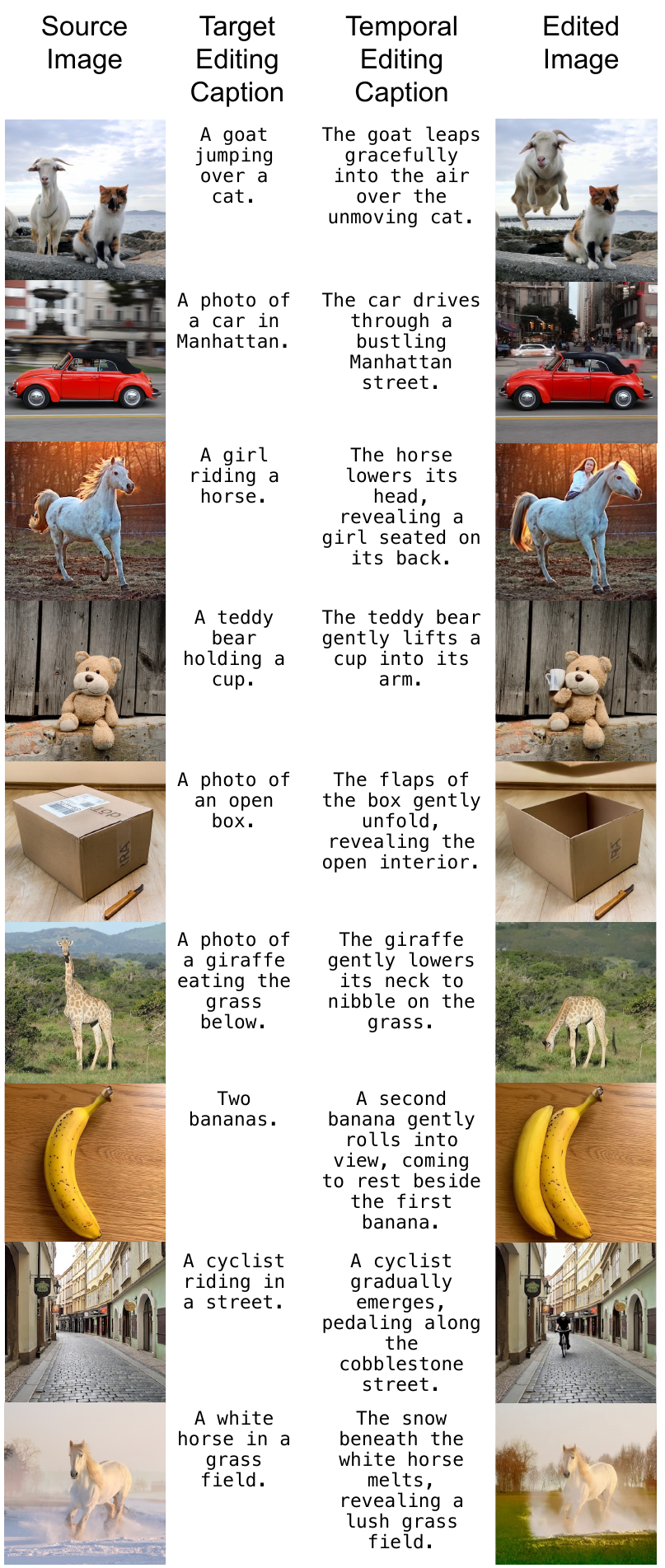}
\caption{\textbf{TEdBench Editing Examples.}}
\label{fig:f2f_examples_2}
\end{figure}

\section{Temporal Editing Captions}\label{supp:temporal_captions}
\subsection{VLM Instruction}
As outlined in \Cref{temporal_captions}, we propose a framework for automatically generating the temporal editing caption by leveraging the original target editing prompt in conjunction with the source image. 
The instruction given to the VLM, along with the source image is:

\textit{
    'Write a one-sentence description of a short video that begins with the provided image and smoothly transitions into a scene of a "CAPTION",
    highlighting how elements in the image undergo changes or movement over time.
    Keep the description simple, concise and short, focusing only on essential changes and actions without altering unnecessary details.
    Avoid mentioning elements that do not contribute to the main change needed, and focus the description on the main transitions.
    Do not add objects that are not in the original image or described in the final scene.
    The camera should remain static unless movement is absolutely necessary.
    Ensure all transitions happen within a few second duration without mentioning the length or using the word "video".'
}

\begin{figure}[H]
\centering
\includegraphics[width=1.03\linewidth]{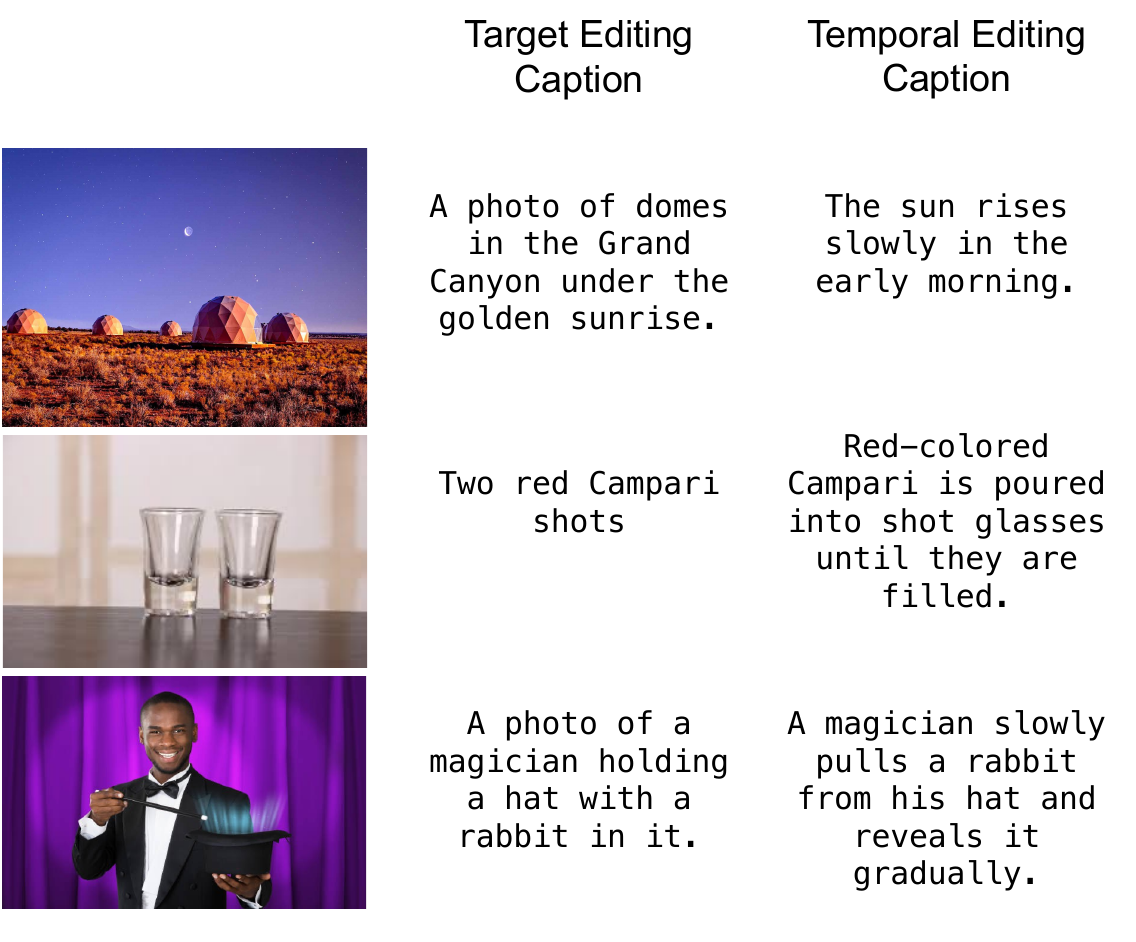}
\caption{\textbf{In Context Learning Examples.}}
\label{fig:icl}
\end{figure}

Here, "CAPTION" is replaced with the target caption specific to the image.
Additionally, as explained, in-context learning is employed to provide the VLM with examples alongside the instruction. Before processing the desired source image and edit prompt, the instruction is presented to the VLM nine times, each paired with a distinct example consisting of a source image, target caption, and corresponding temporal editing caption.
Examples of these are illustrated in \Cref{fig:icl}.

\subsection{Ablation}
To assess the impact of the Temporal Editing Caption, we conduct an ablation experiment comparing its use against directly using the target editing captions from the TEdBench benchmark. 
Apart from this modification, we adhere to the same protocols as described in the original experiment in \Cref{sec:tedbench_results}.
As shown in \Cref{tab:tedench_results_ablation}, this setup preserves a similar resemblance to the source image but underperforms in terms of image editing performance.

\begin{table}[H]
    \centering
    \begin{tabular}{l|cc|c}
        \toprule
        & \multicolumn{2}{c|}{\textbf{Source}} & \textbf{Target} \\
        \raisebox{1.2em}{Model} & 
        \raisebox{1.2em}{LPIPS$_\downarrow$} & 
        \raisebox{1.2em}{CLIP-I$_\uparrow$} & 
        \raisebox{1.2em}{CLIP$_\uparrow$} \\
        \noalign{\vskip -0.6em}
        \midrule
        Original Captions     & \textbf{0.21} & \textbf{0.89} & 0.60  \\
        Temporal Captions    & 0.22 &  \textbf{0.89} & \textbf{0.63} \\
        \bottomrule
    \end{tabular}
    \caption{
        \textbf{Temporal Editing Captions Ablation.}
    }
\label{tab:tedench_results_ablation}
\end{table}

\section{Frame Selection}\label{supp:frame_seletion}
\subsection{VLM Instruction}
As detailed in \Cref{sec:frame_selection}, our method selects the frame that best aligns with the intended edit from each generated video.
To automate this process, inspired by \cite{kim2024image}, we create a collage of uniformly sampled frames from the video, along with the source image and target editing caption, and prompt a VLM to identify the optimal frame.
The model is instructed to select the earliest frame (i.e., with the lowest index) that satisfies the editing intent, minimizing deviation from the original image.
In both this process and the best seed selection process (applied across all methods), if none of the edited frames successfully fulfill the desired edit, the original image is retained as the final output.

The instruction provided to the VLM is:

\textit{
    `The image displays the source photo at the top, with a collage of 12 edited versions beneath it.
    The target edit image caption was: ``CAPTION".
    Your task is to choose the image from 1 to 12 that best follows this edit fully and naturally.
    If none of the images follows the edit, select image 0.
    If multiple images follow the edit equally, prioritize the one with the lowest number possible.
    Avoid selecting images that appear to follow the edit but are not edits of the original image.
    Additionally, avoid images where camera motion, zoom, or image quality differs significantly, or where the content does not appear stable relative to the original source.
    Respond with: ``The selected edit is:x" where x is the number of your chosen edit.'
}

\begin{figure}[H]
\centering
\includegraphics[width=1\linewidth]{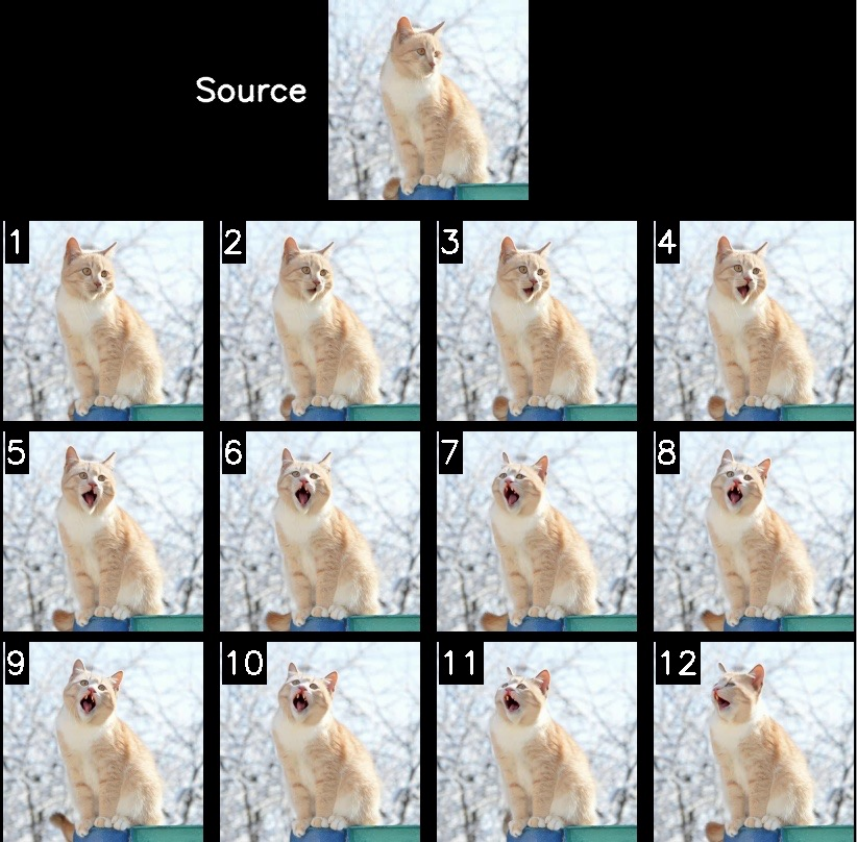}
\caption{\textbf{Frame Selection Collage.} The target editing caption for this example is: ``A photo of a cat yawning.".}
\label{fig:frame_collage}
\end{figure}

Here, "CAPTION" refers to the target editing caption. Examples of the collages can be found in \Cref{fig:frame_collage}.

\subsection{Ablation}
To validate the effectiveness of our approach, we compare it to the naive solution of using the last frame of the generated video as the edited output.
The evaluation follows the same protocol described in \Cref{sec:tedbench_results}.
As can be seen in \Cref{tab:tedench_frame_ablation}, this naive approach results in a lower target CLIP score for the edited outputs, highlighting the advantages of our method.

\begin{table}[H]
    \centering
    \begin{tabular}{l|cc|c}
        \toprule
        & \multicolumn{2}{c|}{\textbf{Source}} & \textbf{Target} \\
        \raisebox{1.2em}{Model} & 
        \raisebox{1.2em}{LPIPS$_\downarrow$} & 
        \raisebox{1.2em}{CLIP-I$_\uparrow$} & 
        \raisebox{1.2em}{CLIP$_\uparrow$} \\
        \noalign{\vskip -0.6em}
        \midrule
        Last Frame     & 0.24 & \textbf{0.9} & 0.61  \\
        Selected Frame   & \textbf{0.22} &  0.89 & \textbf{0.63} \\
        \bottomrule
    \end{tabular}
    \caption{
        \textbf{Frame Selection Ablation.}
    }
\label{tab:tedench_frame_ablation}
\end{table}

\section{Editing Manifold Pathway}\label{supp:manifold}
\begin{figure}[H]
\centering
\includegraphics[width=0.85\linewidth]{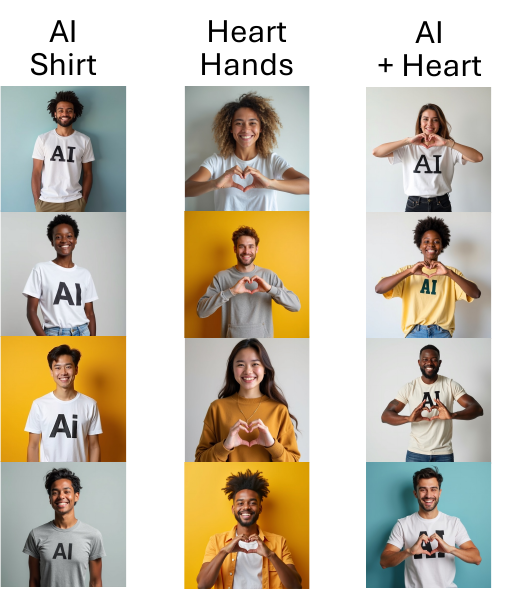}
\caption{\textbf{Flux.1-dev Generations}}
\label{fig:flux}
\end{figure}

As elaborated in \Cref{sec:manifold}, to simulate the image manifold, we generated 200 images across three distinct categories using FLUX.1-dev.
Examples of these generated images are shown in~\Cref{fig:flux}.

\section{Inference Hyperparameters}
We use the 5B-parameter image-to-video version of CogVideoX, ensuring consistency by applying the same inference hyperparameters across all experiments.
The model generates a fixed 49 frames per sample. During inference, we use the default denoising scheduler in CogVideoX, which is based on DDIM with V-prediction. We perform 50 denoising steps and set the classifier-free guidance scale for text conditioning to 6.0.

\section{PosEdit}\label{supp:posedit}
\begin{figure}[H]
\centering
\includegraphics[width=0.86\linewidth]{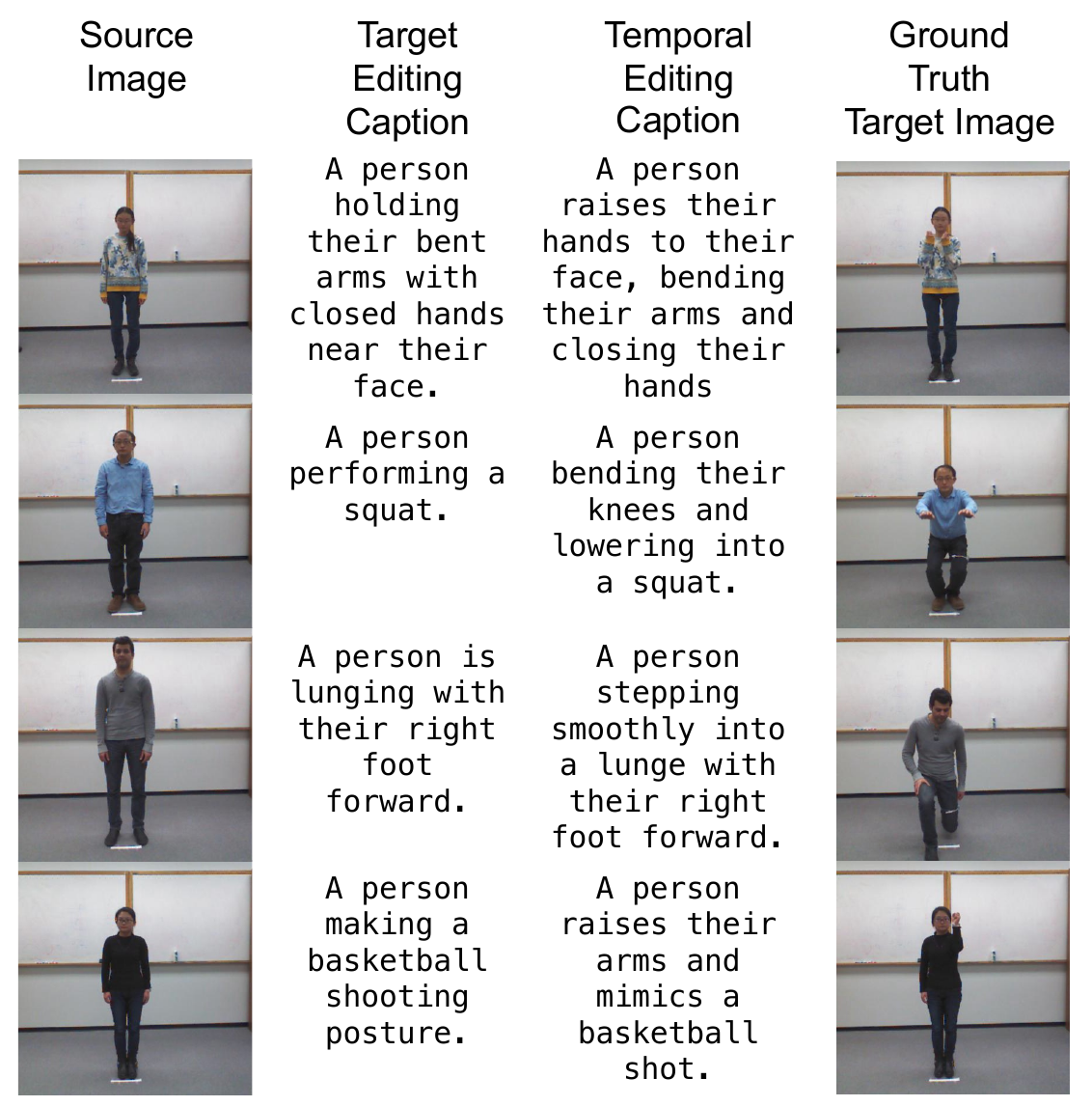}
    \caption{\textbf{PosEdit Examples.}}
\label{fig:davidata_examples_1}
\end{figure}

\begin{figure}[H]
\centering
\includegraphics[width=0.86\linewidth]{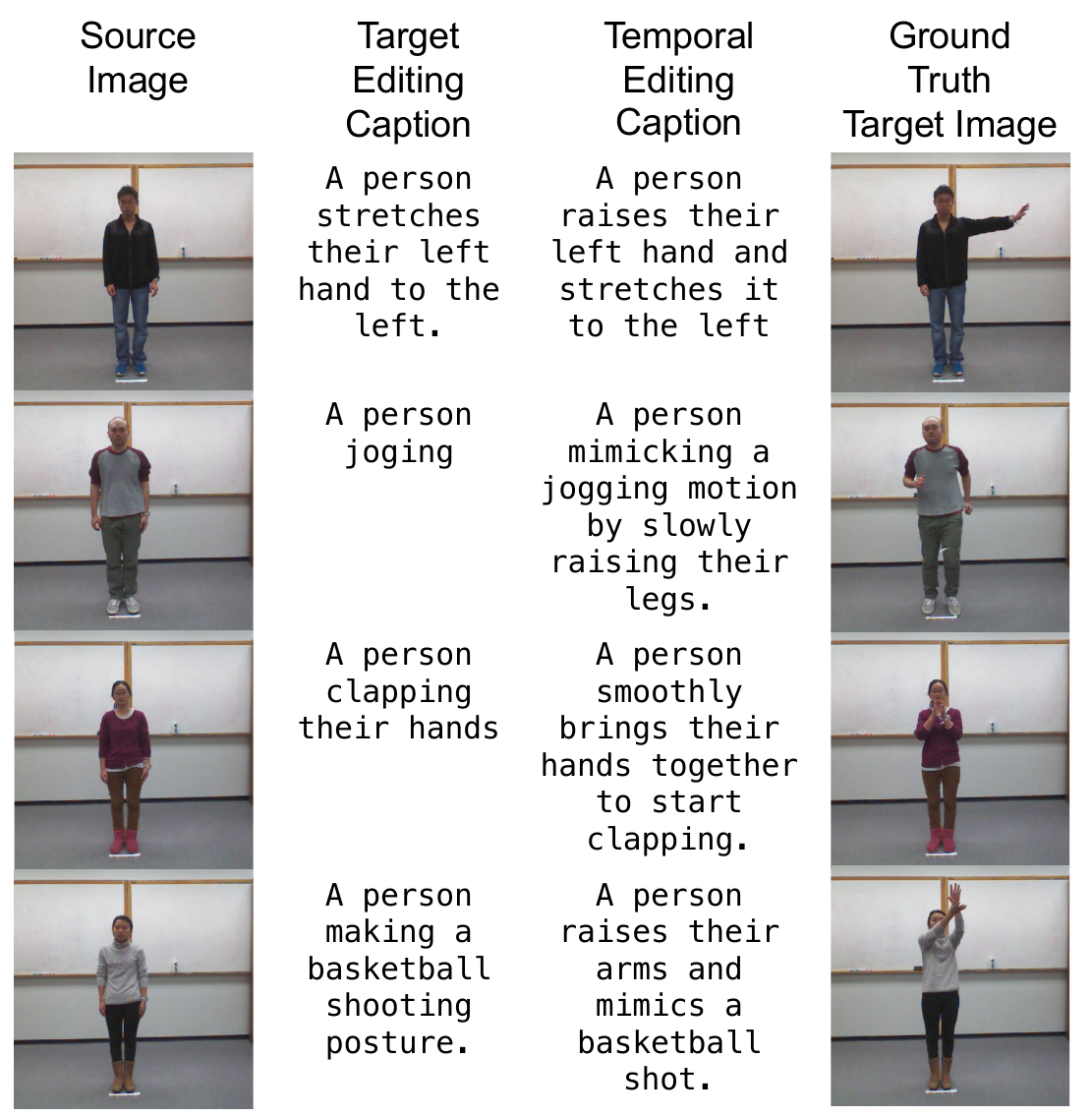}
\vspace{-5pt}
\caption{\textbf{PosEdit Examples.}}
\label{fig:davidata_examples_2}
\vspace{-5pt}
\end{figure}

\begin{figure}[H]
\centering
\includegraphics[width=1\linewidth]{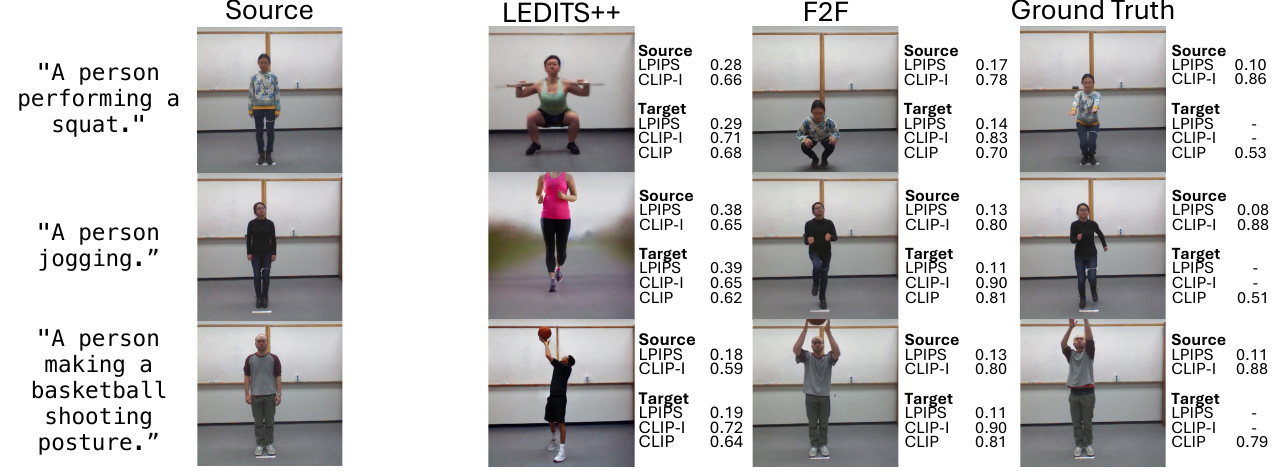}
\vspace{-5pt}
\caption{\textbf{PosEdit Quantitative Evaluation Examples.}}
\label{fig:davidata_examples_metrics}
\vspace{-5pt}
\end{figure}

\vspace{-2pt}
In \Cref{sec:posedit}, we outline the construction of a human pose editing dataset.
This dataset encompasses 58 editing tasks, distributed across 8 distinct action categories, featuring 8 different subjects.
The source images consistently depict a neutral standing pose with arms relaxed at the sides, while the target poses vary according to the edit category.
Each editing category is paired with a target caption and a temporal caption.
\Cref{fig:davidata_examples_1} and \Cref{fig:davidata_examples_2} illustrates examples for each action category.
Additionally, in \Cref{fig:davidata_examples_metrics}, we complement \Cref{fig:posedit_qualitative} with quantitative results, demonstrating how the numerical evaluation aligns with its intended measurement objectives.

\vspace{-5pt}
\section{Human Survey}\label{supp:human_survey}
\begin{figure}[H]
\centering
\includegraphics[width=1.0\linewidth]{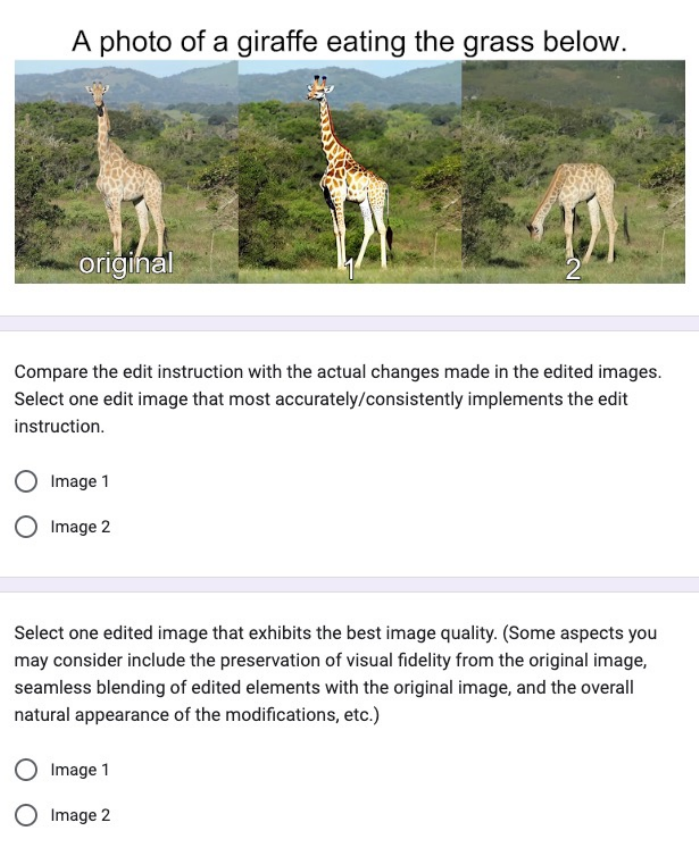}
\caption{\textbf{Survey Example.}}
\label{fig:survey_1}
\end{figure}

\begin{figure}[H]
\centering
\includegraphics[width=1.0\linewidth]{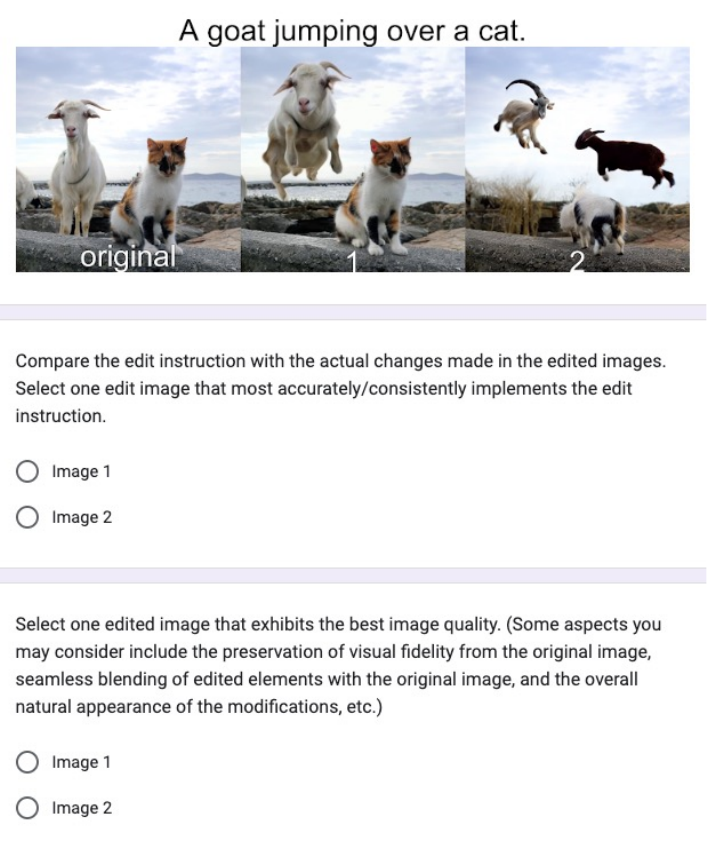}
\caption{\textbf{Survey Example.}}
\label{fig:survey_2}
\end{figure}

\vspace{-5pt}
As detailed in \Cref{sec:survey}, we conducted a human evaluation survey to assess our method's performance based on real user preferences. 
Following the framework in \cite{zhang2024magicbrush}, the survey questions evaluated (1) the accuracy of the edit relative to the prompt and (2) the quality of the edit, defined as the preservation of visual fidelity to the source image.
Each participant reviewed 20 edits, comparing our method with LEDITS++.
Examples of the pages shown to the evaluators are provided in Figures \ref{fig:survey_1} and \ref{fig:survey_2}.

\section{Further Ablations}\label{supp:more_ablations}
We present additional ablation results demonstrating the ability of our method to handle two distinct editing challenges: (1) diverse backgrounds and (2) out-of-video-distribution edits.

\noindent \textbf{Backgrounds.} Similar to our image generation approach in \Cref{sec:manifold}, we generate eight images of two objects with varying backgrounds.
Each image generation prompt follows one of the two templates below, where LOCATION is replaced with one of the following settings:
colorful amusement park,
sandy beach,
magical fairy tale forest,
futuristic cityscape,
old library,
snowy mountain peak,
bustling train station,
and quaint village square.
\begin{enumerate}
    \item \textit{``A small brown teddy bear sitting in} LOCATION\textit{.''}
    \item \textit{``A tennis ball lies in} LOCATION\textit{.''}
\end{enumerate}

\noindent \textbf{Out-of-video-distribution editing.} As discussed in \Cref{sec:limitations}, one might assume that our method would fail to perform edits requiring temporal transformations that deviate significantly from typical real-world videos (the model's training set).
To show that this is not necessarily the case, we selected two unreal editing processes, using the following temporal captions:
\begin{enumerate}
    \item \textit{``The teddy bear slowly rips, revealing stuffing coming out.''}
    \item \textit{``The tennis ball gradually transforms into a ripe red tomato.''}
\end{enumerate}

\begin{figure}[H]
\centering
\includegraphics[width=\linewidth]{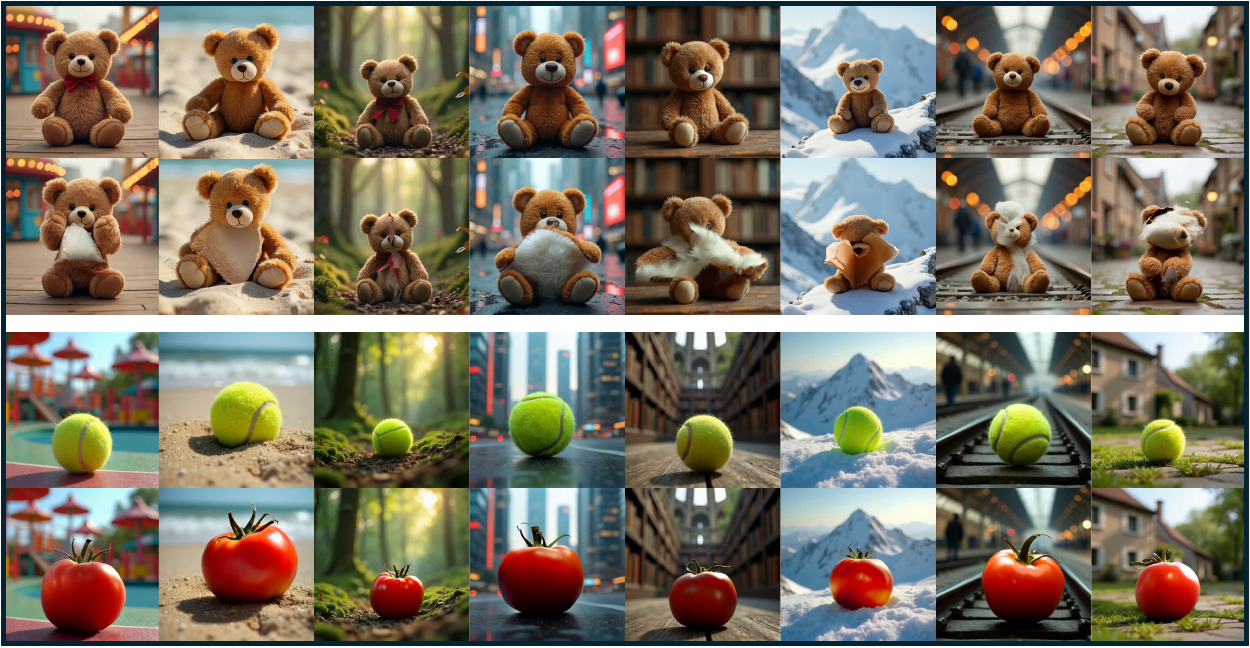}
\caption{\textbf{Ablation Examples.}}
\label{fig:tennis_tomato_backgrounds}
\end{figure}

These edits depict unrealistic events that do not ordinarily occur in real-life footage. However, as shown in \cref{fig:tennis_tomato_backgrounds}, our method successfully handles both the background variations and the out-of-distribution nature of these transformations. Moreover, the full temporal sequences in \cref{fig:f2f_videos} demonstrate that these edits occur seamlessly without any external interaction (the teddy bear rips apart spontaneously, and the tennis ball morphs magically into a tomato).

\section{Additional Vision Tasks Captions}\label{supp:vision_tasks}
As outlined in \Cref{sec:silver_tasks} and demonstrated in \Cref{fig:generatior_example}, we showcase our framework's applicability for additional, more classic vision tasks that are not typically classified as image editing. 
For these tasks, we employ Runway Gen-3 as our video generator. Empirically, these tasks required longer and more descriptive captions. 
The temporal editing captions used for each task are as follows:
\begin{enumerate}
\item \textbf{Relighting}:
 \textit{
 'The scene's lighting shifts gradually, changing to night. The sun is setting, and artificial lights replace it. 
 The camera is static. Time-lapse. Cinematic.'
}

\item \textbf{Outpainting}:
 \textit{
'The image expands, adding new surroundings seamlessly beyond the original frame.'
}

\item \textbf{Denoising}:
 \textit{
'The image clears up as noise fades away, revealing smoother, cleaner details.'
}

\item \textbf{Debluring}:
 \textit{
'The camera comes into focus, revealing sharp details and enhanced clarity, as though a camera lens has adjusted perfectly. Nothing moves. Static image.'
}

\end{enumerate}